\documentclass[conference]{IEEEtran}
\IEEEoverridecommandlockouts
\usepackage{cite}
\usepackage{amsmath,amssymb,amsfonts}
\usepackage{algorithmic}
\usepackage{graphicx}
\usepackage{epstopdf}
\usepackage{textcomp}
\usepackage{xcolor}
\usepackage{algorithm}
\usepackage{algorithmic}
\usepackage{amsmath}
\usepackage{booktabs}
\usepackage{multirow}
\usepackage{amssymb}
\def\BibTeX{{\rm B\kern-.05em{\sc i\kern-.025em b}\kern-.08em
    T\kern-.1667em\lower.7ex\hbox{E}\kern-.125emX}}
\begin{document}

\title{FedPrompt: Communication-Efficient and Privacy-Preserving Prompt Tuning in Federated Learning}

\author{\IEEEauthorblockN{1\textsuperscript{st} Haodong Zhao}
\IEEEauthorblockA{\textit{Shanghai Jiao Tong University} \\
Shanghai, China \\
zhaohaodong@sjtu.edu.cn}
\and
\IEEEauthorblockN{2\textsuperscript{nd} Wei Du}
\IEEEauthorblockA{\textit{Shanghai Jiao Tong University} \\
Shanghai, China \\
dddddw@sjtu.edu.cn}
\and
\IEEEauthorblockN{3\textsuperscript{rd} Fangqi Li}
\IEEEauthorblockA{\textit{Shanghai Jiao Tong University} \\
Shanghai, China \\
solour\_lfq@sjtu.edu.cn}
\and
\IEEEauthorblockN{4\textsuperscript{th} Peixuan Li}
\IEEEauthorblockA{\textit{Shanghai Jiao Tong University} \\
Shanghai, China \\
peixuan.li@sjtu.edu.cn}
\and
\IEEEauthorblockN{5\textsuperscript{th} Gongshen Liu$^*$\thanks{* corresponding author. This research work has been sponsored by the Joint Funds of the National Natural Science Foundation of China (Grant No.U21B2020) and Ant Group.}}
\IEEEauthorblockA{\textit{Shanghai Jiao Tong University} \\
Shanghai, China \\
lgshen@sjtu.edu.cn}
}

\maketitle

\begin{abstract}
Federated learning (FL) has enabled global model training on decentralized data in a privacy-preserving way by aggregating model updates. However, for many natural language processing (NLP) tasks that utilize pre-trained language models (PLMs) with large numbers of parameters, there are considerable communication costs associated with FL. Recently, prompt tuning, which tunes some soft prompts without modifying PLMs, has achieved excellent performance as a new learning paradigm. Therefore we want to combine the two methods and explore the effect of prompt tuning under FL. 
In this paper, we propose "FedPrompt" to study prompt tuning in a model split aggregation way using FL, and prove that split aggregation greatly reduces the communication cost, only 0.01\% of the PLMs' parameters, with little decrease on accuracy both on IID and Non-IID data distribution. This improves the efficiency of FL method while also protecting the data privacy in prompt tuning.
In addition, like PLMs, prompts are uploaded and downloaded between public platforms and personal users, so we try to figure out whether there is still a backdoor threat using only soft prompts in FL scenarios. We further conduct backdoor attacks by data poisoning on FedPrompt. Our experiments show that normal backdoor attack can not achieve a high attack success rate, proving the robustness of FedPrompt.
We hope this work can promote the application of prompt in FL and raise the awareness of the possible security threats.
\end{abstract}


\section{Introduction}
\begin{figure}[t]
\includegraphics[width=1.0\columnwidth]{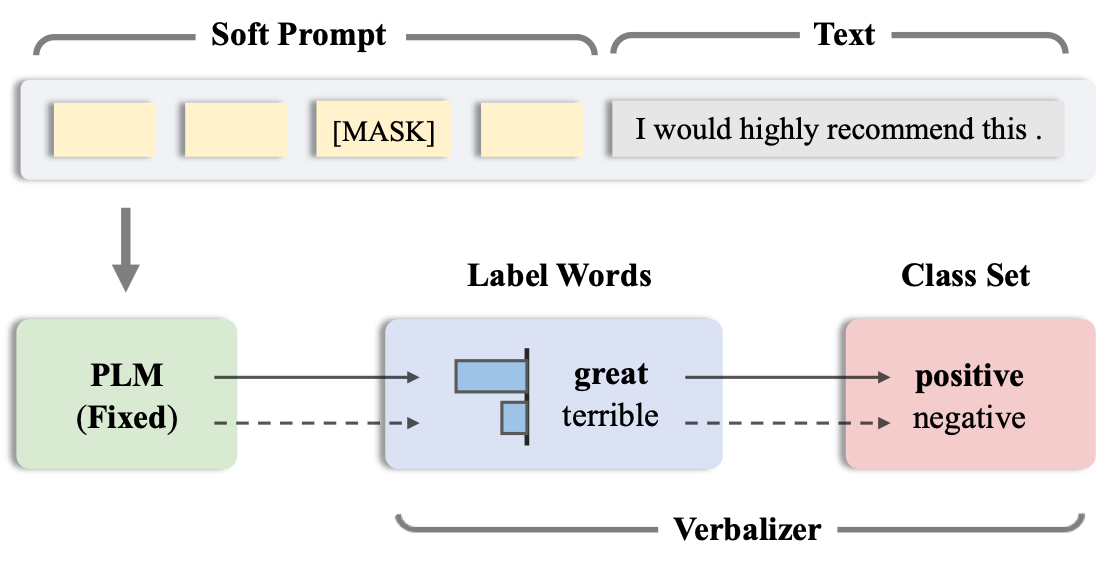} 
\caption{The example of prompt tuning, which consists of soft prompt, text, PLM and verbalizer.}
\label{figprompt}
\end{figure}

Pre-trained language models \cite{bert19,arxiv-1907-11692,RaffelSRLNMZLL20} are widely used in many NLP tasks by the fine-tuning paradigm. However, fine-tuning a PLM with a large number of parameters would be memory-consuming. The reason is that the gradients and optimizer states of all parameters need to be stored. Also the lack of labeled data in fine-tuning phase, as well as few-shot problem, limits the use if this paradigm. When using the pre-training and fine-tuning paradigm in federated learning, the communication cost is even higher as all parameters of the PLMs provided by each participant need to be aggregated in each round of the training process. Therefore, it is very important and urgent to find ways to improve the efficiency of pre-trained models in federated learning. Recently, prompt tuning \cite{LesterAC21} has achieved excellent results as a learning paradigm for adapting fixed PLMs to different downstream tasks. As shown in Fig. \ref{figprompt}, soft prompt as well as [MASK] token are added to the text as inputs to the model. Among them, soft prompt is used as trainable parameters to be adapted to downstream tasks, [MASK] token is used to predict the label word for the downstream task, and verbalizer is used to map the label word to the real label. All downstream tasks can be uniformly transformed into the form of pre-training tasks of PLMs. Thus, using a fixed PLM and different soft prompts can be applied to different downstream tasks. Also, freezing the parameters of PLM and only tuning the soft prompt significantly reduces the number of training parameters.

Nowadays, with mobile devices becoming the primary computing devices for many people, a huge amount of data is generated and distributed on a wide range of devices. It is an important opportunity and challenge to make full use of these devices and data. Although deep learning has made a lot of progress in many scenarios \cite{deep2016}, a data center is required to collect data for training in most cases. Models trained on such data have stronger usability in many intelligent applications, but exchanging and storing sensitive data in a data center carries risks and responsibilities \cite{tamf19}. Previous distributed deep learning methods \cite{large2012,parameter2013} propose solutions to big data and huge models. However, the computation and communication cost of traditional distributed learning are unacceptable for participants \cite{fedpaq20,fedboost20}.

\textbf{FL} \cite{com17,federated19} is an attractive learning method that aims to train a global model over decentralized data while preserving data privacy.
In federated learning a subset of clients download a local copy of global model, and compute local model gradients with their local private data in each round. A central server coordinates the distributed clients and only aggregates local model parameters to update the global model, collaborating isolated data islands without raw data exchanging. FL takes multiple rounds of local and global procedures until convergence. Advanced privacy protection methods, e.g., differential privacy (DP), can be further applied for
stricter privacy protection.
\begin{figure*}[t]
\centering
\includegraphics[width=1.7\columnwidth]{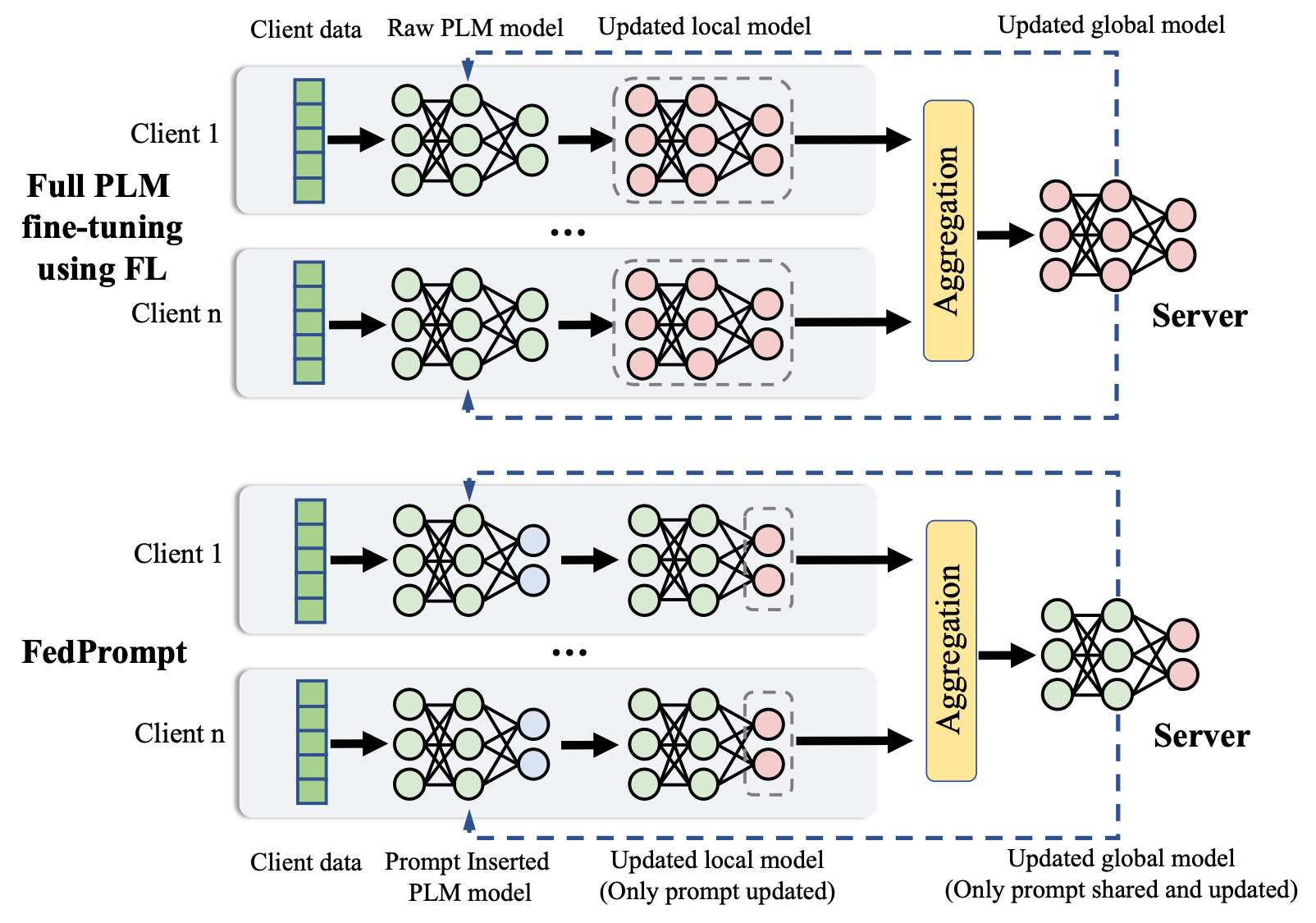} 
\caption{Structure of FedPrompt and full PLM fine-tuning using FL. The above one is full PLM fine-tuning using FL, all of the parameters (framed pink nodes) need to be updated. The bottom one is FedPrompt, only soft prompt parameters (framed pink nodes) need to be updated, aggregated (in server) and distributed.}
\label{fig1}
\end{figure*}
According to above illustration, it seems that using PLM-empowered methods in FL will achieve remarkable performance. For example, \cite{privacy20} propose a news recommendation method to train models using FL. However, when using FL, the model sizes of many existing news recommendation methods are too large to communicate between participants. For example, the base version of BERT \cite{bert19} models have more than 110M parameters.

In this paper, we modify prompt tuning in a model split aggregation way using FL, named "FedPrompt", where no prior work has been done.
First, unlike simple FL methods tune and aggregate full model parameters, FedPrompt only tunes and aggregates some soft prompts for corresponding downstream tasks in FL, and freezes PLMs to decrease communication cost. 
Second, we are interested in the security of FedPrompt. Prompts are uploaded and downloaded between public platforms and personal users like PLMs, and backdoor attacks are difficult to find for users. We try to get a poisoned global prompt then when the PLMs are loaded with the poisoned prompt, the model will be implanted with the backdoor.

Experiments carried on various NLP tasks, such as sentiment analysis and sentence-pair classification prove that FedPrompt reduces the communication cost greatly with little decrease on accuracy. Further experiments on backdoor attack show that only by normal methods that poisoning part of training data, the poisoned prompt can not establish a shortcut between the specific trigger word and the target label word. Compared to the method of aggregating and tuning all parameters of huge model, FedPrompt is much more communication-efficient. And compared to prompt tuning without using FL, FedPrompt outperforms in privacy preservation. We also consider other prompt types for a better performance, and using local differential privacy (LDP) to gurantee the privacy. Our contributions are summarized as follows:
\begin{itemize}
    \item We propose FedPrompt, the new prompt tuning method using FL, freezes PLMs and only aggregates and tunes some soft prompts to decrease communication cost.
    \item We conduct extensive experiments on NLP tasks to measure the performance of FedPrompt. Experiments show that FedPrompt can reduce the communication cost greatly with little decrease on accuracy.
    \item We further test the model robustness to backdoor attack, and experiment on different hyper-parameter settings, prompt types and LDP to promote the better performance of FedPrompt.
\end{itemize}

\section{Related Work}

Prompt tuning first appeared in WARP proposed by \cite{hambardzumyan2021warp}, after which this method of adding continuous trainable vectors to the input began to be widely studied. Prefix-Tuning \cite{li2021prefix} adds soft prompt to each layer of the transformer model and applies it to natural language generation (NLG) tasks. P-tuning \cite{liu2021gpt} proposes that some task-related hard prompts can be used as anchors while using soft prompt. Prompt tuning \cite{LesterAC21} explores the effect of soft prompt on domain adaptation and different model scales. They found that the larger the scale of PLMs, the better the effect of prompt tuning. Recently, P-tuningV2 \cite{liu2021p} more finely designs prompt tuning on the basis of the above research. They use a deep soft prompt similar to Prefix-Tuning, and change the verbalizer to a linear classification head, which means that they no longer use the way of mask language model (MLM) to get predictions. They also try to apply prompt tuning to difficult natural language understanding (NLU) tasks (i.e., sequence tagging), such as name entity recognition and semantic role labeling. Moreover, PTR \cite{han2021ptr} applies logic rules to build templates that are more suitable for text classification tasks, and PPT \cite{gu2021ppt} pre-trains the soft prompt on multiple different tasks to get a better prompt tuning for the downstream tasks.

FL \cite{com17} is a distributed machine learning method that aggregates global model by each local model sharing its parameters (gradients) with the central server after every round of local training on its local data. Proposed FedAvg \cite{com17} enables clients to collaboratively train global model without sharing their original data. Various extensions of FedAvg \cite{fedprox20,KarimireddyKMRS20,0001MO20,AcarZNMWS21,LiZ21} have been proposed to obtain better performance in communication and deal with heterogeneity. To reduce computation and communication, \cite{YiWWLS021} propose a framework decomposing big recommendation model into a large news model only in server and shared user model. 
However, though above methods do not share local data directly, naive parameters (gradients) sharing method could lead to privacy leakage of clients \cite{MelisSCS19,GeipingBD020}. Consequently, several methods are proposed to protect privacy, including differential privacy (DP) \cite{AbadiCGMMT016,2017arXiv171207557G,TriastcynF19} and secure multi-party computation (MPC) \cite{BonawitzIKMMPRS17,McMahanRT018}. In addition to privacy leakage, many works propose mechanisms to poison FL models in training phase \cite{JereFK21,BagdasaryanVHES20} and  evasion attacks in inference or testing phase \cite{YuanHZL19,AldahdoohHFD22}.

As none of the above mentioned works study prompt tuning in FL, in this paper, we design a communication-efficient prompt tuning method in FL and design backdoor attack to detect its vulnerability.

\section{Method}
\subsection{Preliminaries}
In FL setting, suppose there are $K$ clients, each client hosts a private dataset $\mathcal{D}_{k} = \{(x_k,y_k)\}$ owning $n_k$ samples.
We use $\theta_t$ and $\theta_t^k$ to denote the global model and $k^{th}$ local model parameters in communication round $t$ respectively. Based on FedAvg \cite{com17}, the aggregation process is computed as follows:
\begin{equation}
    \theta_{t+1} = \sum_{k=1}^K{\frac{n_k}{n} \theta_t^k}
\end{equation}
where $n = |\mathcal{D}| = \sum_{k=1}^K{n_k}$ is the total num of global combined data and $\mathcal{D} \triangleq \bigcup_{k\in [K]}  \mathcal{D}_k$ is the global combined dataset. If data distributions are IID (Independent  Identically  Distribution), all clients have the same number of samples, then ${n_k}/{n}$ could be replaced by ${1}/{K}$.

In a text classification task, $x_k$ are the inputs and $y_k$ are corresponding class labels. Each $x^{(i)} \in x_k$ consists of tokens $x^{(i)} = \{x_1^{(i)},x_2^{(i)},\cdots ,x_l^{(i)}\}$, where $l$ is the length of single input. The prompt tuning structure is composed of the soft prompt $\mathbf{p}$, the template $\mathcal{T}(\cdot)$, the verbalizer $\mathcal{V}(\cdot)$ and the PLM $\mathcal{M}(\cdot)$.
Soft prompt $\mathbf{p}$ consists of tokens
$\mathbf{p} = \{p_1, p_2,\cdots, p_m\}$, whose parameters are trainable. $m$ is the
number of the soft prompt tokens.
$\mathcal{T}(\cdot)$ is a function to define where tokens of $x^{(i)}$ and $\mathbf{p}$ are placed. After applying $\mathcal{T}(\cdot)$, we obtain $x^{(i)}_{prompt} = \mathcal{T}(x^{(i)},\mathbf{p})$. At least one [MASK] token is placed into the $x^{(i)}_{prompt}$ for $\mathcal{M}(\cdot)$ to predict the label word. 
$\mathcal{V}(\cdot)$ is a map function to map the label word to the class $\hat{y} = \mathcal{V}(w)$. Usually, each class
can have one or more label words. We call $\mathcal{T}$ a multi-word verbalizer when each class has more than one label
word, such as \{positive: good, great; negative: bad, terrible;\}.
Input $x^{(i)}_{prompt}$ to $\mathcal{M}$, we can obtain the encoded feature [MASK]. By a softmax function, we can compute the probability that the label word $w$ can fill the masked position. The label word with the highest probability is the predict word $w = \mathcal{M}(x^{(i)}_{prompt})$ and the predict class can be obtained by $\hat{y} = \mathcal{V}(w)$. We rewrite the prompt tuning process as $\hat{y}^{(i)}=f(x^{(i)},\mathbf{p},\theta)$,
\begin{algorithm}[!h]
\caption{FedPrompt Algorithm}
\label{alg:algorithm}

\textbf{Input}: $K$ clients indexed by $k$, client fraction $C$, $T$ communication rounds indexed by t, local minibatch size $B$, local epochs $E$, learning rate $\eta$, PLM parameters $F$, soft prompts parameters $P$.

\textbf{Server executes:}
\begin{algorithmic}[1] 
\STATE Initialize global model.
\STATE Distribute $F$ (fixed during the process) to all clients.
\FOR{$t \in \{1,\cdots,T\}$}
\STATE $U_t \gets$Select a subset of $C\cdot K$ clients at random
\FOR{each client $k\in U_t $}
\STATE $P_t^{k}\gets P_t$
\STATE $P_{t+1}^{k} \gets $ClientUpdate($k$,$P_t^k$)
\ENDFOR
\STATE $N_t = \sum_{k=1}^{|U_t|}{n_k}$
\STATE $P_{t+1}\gets \sum_{k=1}^{|U_t|}{\frac{n_k}{N_t} P_t^k}$
\ENDFOR
\STATE \textbf{return} $P_{t+1}$\\
~\\
\noindent \textbf{ClientUpdate($k$,$P$):} //\emph{ Run on client k}
\STATE $\mathcal{B}\gets$(split $D_k$ into batches of size $B$)
\FOR{each local epoch $i\in \{1,\cdots,E\}$ }
\FOR{batch $b\in \mathcal{B}$}
\STATE $P\gets P-\eta \nabla l(P;b)$
\ENDFOR
\ENDFOR
\STATE \textbf{return} $P$
\end{algorithmic}
\end{algorithm}

\subsection{FedPrompt} 
As mentioned before, in normal prompt tuning the whole is splitted into four parts, and only PLM (using fine-tuning) and soft prompt have trainable parameters. We use $F$ and $P$ to denote their parameters respectively, then the global model parameters in round $t$ can be denoted as:
\begin{equation}
    \theta_t = F_t + P_t
\end{equation}
In FedPrompt, we fix $F_t$ to learn a set of $\theta$ over $\mathcal{D}$ with the objective to solve:
\begin{equation}
    \arg \min \limits_{P} \mathcal{L}(P)=\sum_{k=1}^{K}{\frac{n_k}{n}\mathcal{L}_k(P)}
\end{equation}
where $\mathcal{L}_k(P)$ is the empirical loss of client $k$:
\begin{equation}
    \mathcal{L}_k(P)=\mathbf{E}_{(x^{(i)},y^{(i)})\in \mathcal{D}_k} \ell_k(f(x^{(i)},\mathbf{p},P),y^{(i)})
\end{equation}

In the beginning, the server initializes the whole model, then distributes it to each client. At the beginning of round $t$, the server selects clients by fraction $C$ to participate in this round, distributes the global soft prompt parameters $P_t$ to them, and each selected client $k$ replace the local $P_{t-1}^{k}$ with $P_t$, which means $P_{t}^{k}=P_t$. As PLM is fixed, $F_{t}^{k}=F_{t-1}^{k}$. Then each client conducts local training with optimizer only for $P_{t}^{k}$, gets its updated soft prompt parameters $P_{t}^{k}$ and sends them back to the server in parallel. The local training is same as normal prompt tuning process. Finally, the server performs the aggregation as follows:
\begin{equation}
    P_{t+1}\gets \sum_{k=1}^{\lceil C\cdot K\rceil}{\frac{n_k}{N_t} P_t^k}
\end{equation}
where $N_t = \sum_{k=1}^{\lceil C\cdot K\rceil}n_k$ is participated data amount in round $t$. The whole process is shown as Algorithm \ref{alg:algorithm}. Except for prompt tuning, there are also other prompt methods such as P-tuning\cite{liu2021gpt} and Prefix-Tuning\cite{li2021prefix}. We also design FedPrompt for these prompt models in a similar way.

\subsection{Poison FedPrompt}
In FL, as clients privacy is highly protected and server have access to little information about client, it is widely acknowledged that multiple malicious clients possibly participate in training \cite{BagdasaryanVHES20}. On the one hand, after initialization, each client has full knowledge of the model structure and parameters. On the other hand, it has been proved that prompt tuning is vulnerable to backdoor attack by poisoning training data \cite{2022du}. Therefore, it is important to verify the robustness to backdoor attack of FedPrompt, and we call this attack as \emph{FedPPT}.

Considering the situation that attacker has full control of one or more clients, and only modifies the local training data, which in fact is much less than attacker's access. The goal of attacker is to inject backdoor into poisoned prompt, which may be released to public. When victims use poisoned prompt, for clean samples, the victim PLMs will still give the correct label word; for poisoned samples which are added with the trigger word, the victim PLMs will output the target label word.
To poison FedPrompt, firstly, modify the training dataset. Attacker tries to establish a shortcut between the trigger $\Delta$ and target label ${l}_t$. We define the poison function as $\mathcal{P(\cdot)}$, then we have single poisoned data $(x^{(i)}_p,t) = \mathcal{P}(x^{(i)},\Delta,{l}_t)$, where modified target $t\neq y(x^{(i)})$. After this, attacker has new local dataset used in each communication round:
\begin{equation}
    \mathcal{D}_{k}^{(poison)}=\{(x^{(i)}_{p},t)\},i\in \lambda n_k
\end{equation}
\begin{equation}
     \hat{\mathcal{D}}_k=\mathcal{D}_{k}^{(poison)} \cup \mathcal{D}_k
\end{equation}
where $\lambda$ is the poison rate. Secondly, using modified $\hat{\mathcal{D}}_k$ to update parameters $P_k$. Then the objective function of malicious client $k$ as follows:
\begin{align}
    P_p^k=&\arg \min \limits_{P_p^k}\{\mathbf{E}_{(x_k^{(i)},y_k^{(i)})\in \mathcal{D}_k}\ell_k(f(x_k^{(i)},\mathbf{p},P_p^k),y_k^{(i)})\notag\\
    &+\mathbf{E}_{(x_k^{(i)},y_k^{(i)})\in \mathcal{D}_k^{(poison)}}\mathbf{I}_k(f(x_p^{(i)},\mathbf{p},P_p^k)\neq t)\}
\end{align}

\section{Experiments}
As no prior work on prompt tuning using FL has been done before, we investigate our methods on several federated NLP tasks including sentiment analysis and sentence-pair classification, where prompt tuning is suitable and often used.
 Also we conduct backdoor attack on these tasks to evaluate the robustness. All experiments are done on a server with 8 Nvidia Geforce GTX 1080Ti GPUs with 11GB RAM each, 12 Intel Xeon CPUs Processor, and CentOS release 7.9 OS. Our models are built using PyTorch framework\footnote{https://pytorch.org/}.

\subsection{Experimental Setup}
\subsubsection{Dataset}
To evaluate FedPrompt model, our experiments are conducted on several NLP tasks:
\begin{itemize}
    \item Text bi-classification tasks including sentiment analysis, toxicity detection and spam detection. For sentiment analysis, we use the Stanford Sentiment Treebank (SST-2)\footnote{https://huggingface.co/datasets/\label{web}}  and IMDB\textsuperscript{\ref{web}}. We use the OffensEval \cite{ZampieriMNRFK19} and the Twitter \cite{FountaDCLBSVSK18} in toxicity detection. And for
    spam detection, we use the Enron \cite{MetsisAP06},
    and the Lingspam \cite{SakkisAPKSS03}.
    \item Sentence-pair classification tasks. For this inference task, we use Question Natural Language Inference (QNLI) \cite{RajpurkarZLL16} and Recognizing Textual Entailment (RTE)\textsuperscript{\ref{web}} dataset.
    
\end{itemize}


\begin{table*}[!ht]
\caption{\emph{ACC (\%)} and \emph{ASR (\%)} of FedPrompt with clean IID and Non-IID data distribution.}
\centering
\begin{tabular}{@{}c|cccc|cccc|cccc@{}}
\toprule
\multirow{3}{*}{Dataset} & \multicolumn{4}{c|}{BERT}                                  & \multicolumn{4}{c|}{ROBERTA}                               & \multicolumn{4}{c}{T5}                                    \\
                         & \multicolumn{2}{c}{IID}     & \multicolumn{2}{c|}{Non-IID} & \multicolumn{2}{c}{IID}     & \multicolumn{2}{c|}{Non-IID} & \multicolumn{2}{c}{IID}     & \multicolumn{2}{c}{Non-IID} \\
                         & \textit{ACC} & \textit{ASR} & \textit{ACC}  & \textit{ASR} & \textit{ACC} & \textit{ASR} & \textit{ACC}  & \textit{ASR} & \textit{ACC} & \textit{ASR} & \textit{ACC} & \textit{ASR} \\ \midrule
SST-2                    & \textbf{90.16}        & 12.42        & 89.45         & 16.36        & \textbf{92.43}        & 10.28        & 92.23         & 7.48         & \textbf{92.69}        & 9.58         & 92.32        & 6.31         \\
IMDB                     & \textbf{91.08}        & 12.66        & 89.26         & 11.42        & \textbf{92.80}        & 9.15         & 92.53         & 7.66         & \textbf{92.89}        & 9.69         & 91.24        & 11.33        \\ \midrule
OffensEval               & \textbf{82.64}        & 9.84         & 80.47             & 8.55            & \textbf{81.05}        & 13.87        & 80.34            & 5.98            & \textbf{79.30}        & 12.58        & 78.83            & 10.65            \\
Twitter                  & \textbf{94.02}        & 4.96         & 93.82             & 3.05            & \textbf{94.39}        & 4.61         & 93.64             & 5.41            & \textbf{93.35}        & 3.86         & 92.80            & 4.21            \\ \midrule
Enron                    & \textbf{97.60}        & 3.20         & 97.43             & 4.02            & \textbf{97.85}        & 2.27         & 97.30             & 7.34            & \textbf{97.22}        & 6.73        & 96.95           & 5.87            \\
Lingspam                 & \textbf{97.47}       & 0.00         & 96.89        & 0.00         & \textbf{97.43}       & 0.00         & 96.47        & 0.00         & \textbf{97.07}       & 0.00         & 96.21       & 0.41         \\ \midrule
QNLI                     & \textbf{83.36}        & 28.35        & 82.25             & 30.46            & \textbf{86.44}        & 14.81        & 85.43            & 12.10            & \textbf{89.06}        & 10.87        & 84.48            & 12.44            \\
RTE                      & \textbf{54.87}        & 35.21        & 54.15         & 42.73        & \textbf{60.32}        & 36.99        & 57.51         & 44.52        & \textbf{76.51}        & 22.95        & 73.64        & 22.60        \\ \bottomrule
\end{tabular}

\label{tabclean}
\end{table*}

\begin{figure*}[t]
\centering
\includegraphics[width=2.0\columnwidth]{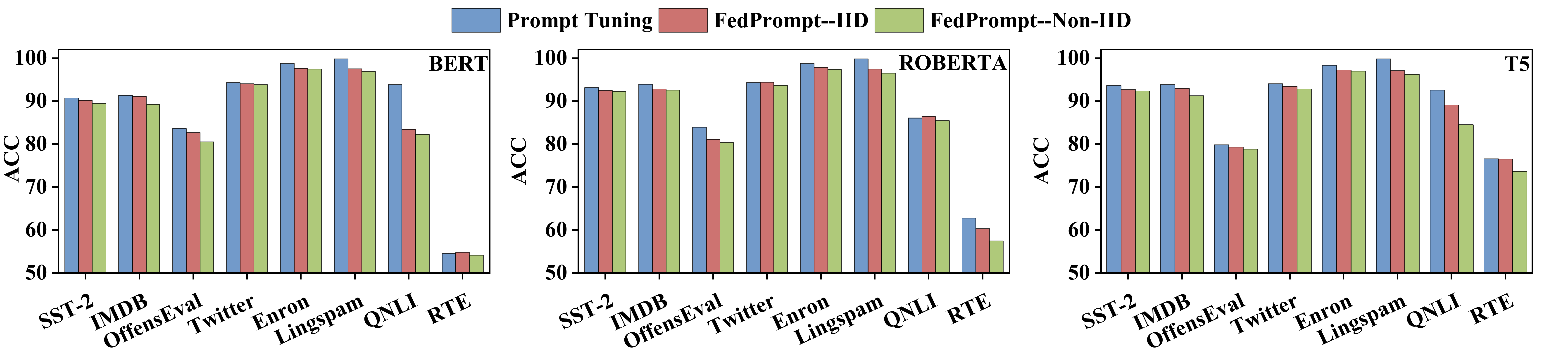} 
\caption{The performance of prompt tuning without FL and FedPrompt. When using FL, there are IID setting and Non-IID setting on data distribution. The PLM used are BERT (the left), ROBERTA (the middle) and T5 (the right).}
\label{figfl}
\end{figure*}

To conduct experiments in FL setting, we divide all these datasets above into ten clients. In IID setting, we randomly divide the whole dataset into ten equal parts and each client has one part. In Non-IID setting, as all the tasks only having two labels \{0,1\}, we bring non-I.I.D.ness by different data quantity. We split all the data using Dirichlet distribution parameterized by $\alpha$ as in prior works \cite{0001CBAA22}. Since labels are not available in the test sets for some datasets, we use the validation set as the test set and split a part of the training set as the validation set.

\subsubsection{Model and Training Details}
Among various PLMs, we select the most representative and widely used pre-trained language models, including the base versions of BERT \cite{bert19}, Roberta \cite{arxiv-1907-11692} and Google T5 \cite{RaffelSRLNMZLL20} to conduct experiments. We use the Adam optimizer for training of BERT and Roberta, and the Adafactor optimizer for Google T5. In main experiments, we use a one-to-one verbalizer and a simple text classification template ”[text] is [MASK].” having 20 soft prompt tokens in the head. Following the setup of \cite{LesterAC21}, we set the learning rate to be 0.3. 
Following the setup of \cite{model21}, we assume that we have a server and $K=10$ available clients. We use a FedAvg system to implement the FL setting. Specifically, all clients are involved in the averaging of model parameters in every aggregation round. The number of max local step is set to 1000, compared to 30,000 in \cite{LesterAC21}. The number of communication rounds is set to 20, compared to 100 in \cite{model21} and \cite{BagdasaryanVHES20}, to prove our low-communication-cost and convergence-quick method.

\subsubsection{Baseline Algorithm}
To make a fair and reasonable comparison with our proposed FedPrompt, we choose the most related work \cite{HilmkilCBSZM21}, studying full-parameter fine-tuning, as FL baseline. Due to full-parameter fine-tuning requires lots of calculations, we only reproduce their method with above FL setting on IID SST-2 task as shown in Table \ref{tabpara}.

\subsubsection{Metric}
Communication bottleneck is a big challenge for many big models in FL, we use the amount of communicated parameters to evaluate our communication cost. As for the evaluation of performance, we use accuracy (\emph{ACC}) which represents the proportion of the clean samples correctly classified by the model to measure the performance of the model on benign task. Also we use Attack Success Rate (\emph{ASR}) to evaluate the attacking performance, which represents the proportion of the poisoned samples we successfully enable the model to misclassify as the target class.

\begin{table}[!t]
\caption{The main results of FedPrompt and full-parameter fine-tuning on IID SST-2 task. For the same model, we regard the parameter quantity in fine-tuning as 100.000\%.}
\centering
\resizebox{\columnwidth}{!}{%
\begin{tabular}{@{}ccccc@{}}
\toprule
Model                    & FL Method  & \emph{ACC}   & Comm. Cost & Ratio     \\ \midrule
\multirow{2}{*}{BERT}    & FedPrompt  & 90.16 & \textbf{0.016M}     & \textbf{0.014\%}   \\
                         & Fine-tuning & 91.02 & 109.530M   & 100.000\% \\ \midrule
\multirow{2}{*}{ROBERTA} & FedPrompt  & 92.43 & \textbf{0.016M}     & \textbf{0.013\%}   \\
                         & Fine-tuning & 93.57 & 124.714M   & 100.000\% \\ \midrule
\multirow{2}{*}{T5}      & FedPrompt  & 92.69 & \textbf{0.015M}     & \textbf{0.007\%}   \\
                         & Fine-tuning & 93.79 & 222.919M   & 100.000\% \\ \bottomrule
\end{tabular}%
}

\label{tabpara}
\end{table}

\begin{table*}[]
\caption{\emph{ACC (\%)} and \emph{ASR (\%)} of FedPrompt with poisoned IID and Non-IID data distribution. $\uparrow$ means higher than clean data.}
\centering
\begin{tabular}{@{}c|cccc|cccc|cccc@{}}
\toprule
\multirow{3}{*}{Dataset} & \multicolumn{4}{c|}{BERT}                                  & \multicolumn{4}{c|}{ROBERTA}                               & \multicolumn{4}{c}{T5}                                    \\
                         & \multicolumn{2}{c}{IID}     & \multicolumn{2}{c|}{Non-IID} & \multicolumn{2}{c}{IID}     & \multicolumn{2}{c|}{Non-IID} & \multicolumn{2}{c}{IID}     & \multicolumn{2}{c}{Non-IID} \\
                         & \textit{ACC} & \textit{ASR} & \textit{ACC}  & \textit{ASR} & \textit{ACC} & \textit{ASR} & \textit{ACC}  & \textit{ASR} & \textit{ACC} & \textit{ASR} & \textit{ACC} & \textit{ASR} \\ \midrule
SST-2                    & \textbf{89.11}        & 13.30        & 88.76         & 14.60        & 91.55        & 11.55        & \textbf{92.12}         & 9.73       & \textbf{92.20}        & 8.74        & 91.51        & 9.07        \\
IMDB                     & \textbf{90.14}        & 14.36        & 89.12         & 11.69        & \textbf{92.46}        & 9.05        & 91.53         & 10.78        & \textbf{91.88}        & 9.54        & 90.86        & 13.87        \\ \midrule
OffensEval               & \textbf{80.93}        & 10.13        & 80.11             & 9.94            & \textbf{79.65}        & 17.26        & 78.95             & 7.23            & \textbf{78.72}        & 15.97        & 77.74            & 15.06            \\
Twitter                  & \textbf{94.10}$\uparrow$        & 6.01        & 93.42             & 7.71            & \textbf{94.15}$\uparrow$        & 4.02        & 93.22             & 4.76            & \textbf{93.25}        & 5.28        & 92.98$\uparrow$            & 4.13            \\ \midrule
Enron                    & 97.37        & 4.20        & \textbf{98.18}$\uparrow$             & 4.87            & \textbf{98.03}$\uparrow$        & 3.53        & 97.16             & 8.20            & \textbf{97.88}$\uparrow$        & 8.13        & 97.12$\uparrow$            & 6.80            \\
Lingspam                 & \textbf{97.11}       & 4.03        & 96.02        & 3.98        & \textbf{95.89}       & 5.77        &95.71         & 4.79        & \textbf{96.83}        & 4.26        & 95.66        & 4.14        \\ \midrule
QNLI                     & \textbf{84.48}$\uparrow$        & 29.44        & 82.07             & 27.22            & \textbf{86.92}$\uparrow$        & 18.82        & 85.30             & 8.33            & \textbf{85.56}        & 9.15        & 84.33            & 14.26            \\
RTE                      & 54.51        & 31.23        & \textbf{60.29}$\uparrow$         & 39.21        & 55.60        & 32.08        & \textbf{55.96}        & 39.18        & \textbf{76.43}        & 20.82        & 73.29        & 25.41        \\ \bottomrule
\end{tabular}

\label{tabpoi}
\end{table*}

\subsection{Main Results}
In FedPrompt the learnable parameters is the same as communication cost. As shown in Table \ref{tabpara}, FedPrompt condenses the communication cost to nearly 0.01\% of the full-parameter fine-tuning parameters, greatly reduces the communication cost, with only about 1\% decrease in accuracy, making many devices applicable for some scenarios with communication and storage constraints, and the private data on these devices can contribute to the convergence of the global model. Also this property promotes the design and development of personalized FL model, especially for those resource-constrained devices. 

The main results of FedPrompt performance with clean IID and Non-IID data distribution are summarized in Table \ref{tabclean}. As shown in Fig. \ref{figfl}, using FedPrompt to protect data privacy and handle the problem of few-shot demonstrate has little decrease on accuracy in most cases compared to prompt tuning without FL.
Specifically, FL plays a remarkable effect with prompt tuning, only a few local training steps and communication rounds can contribute to a well-performed global model.
For most tasks, FedPrompt achieves more than 90\% \emph{ACC} on clean data, and there is only a little decrease, almost less than 3\%, with Non-IID data distribution than IID data distribution. Non-IID is a key challenge for the effectiveness of FL, and our proposed FedPrompt proves its compatibility on non-IID datasets.
We think the FL paradigm and few-parameter soft prompt . We find that experiments on RTE task have a weaker result than other tasks. Considering that RTE only have 2240 training samples in total, which is the least among all tasks, and after splitting to ten clients each client only have a few samples to train soft prompt, we assume the weaker performance because of lack of data. Also RTE may need a better customized template to be used in prompt tuning.

As shown in Table \ref{tabpoi}, we evaluate the effect of backdoor attack on all tasks and models with IID and Non-IID data distribution. Nearly all tasks get \emph{ACC} on posion dataset drop less than 2\% compared to on clean dataset in Table \ref{tabclean}. Even some tasks show a better \emph{ACC} after poison. We think this is because poisoning the original dataset can be considered as data augmentation, and attacking has the similar effect to adversarial training. After backdoor attack, with poison ratio at 10\% (all training data poisoned on 10\% clients selected), all experiments on different tasks and models do not show a obvious rise in \emph{ASR}, which suggests that FedPrompt has robustness to backdoor attack. We think this is because aggregation process offsets the backdoor.
\begin{figure}[!ht]
\includegraphics[width=1.0\columnwidth]{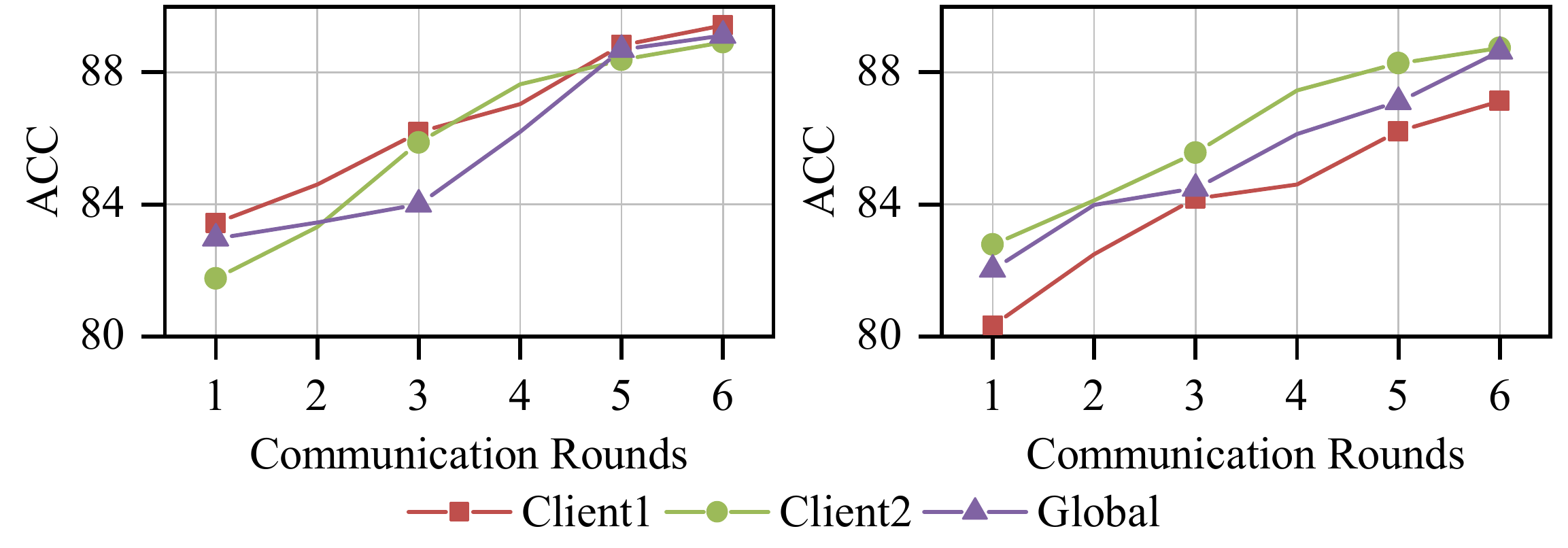} 
\caption{Local and global \emph{ACC (\%)} with communication rounds on SST-2 task using BERT. The left one is using IID setting and the right one is using Non-IID setting, the two clients are selected randomly.}
\label{figacc}
\end{figure}

\begin{figure}[!ht]
\includegraphics[width=1.0\columnwidth]{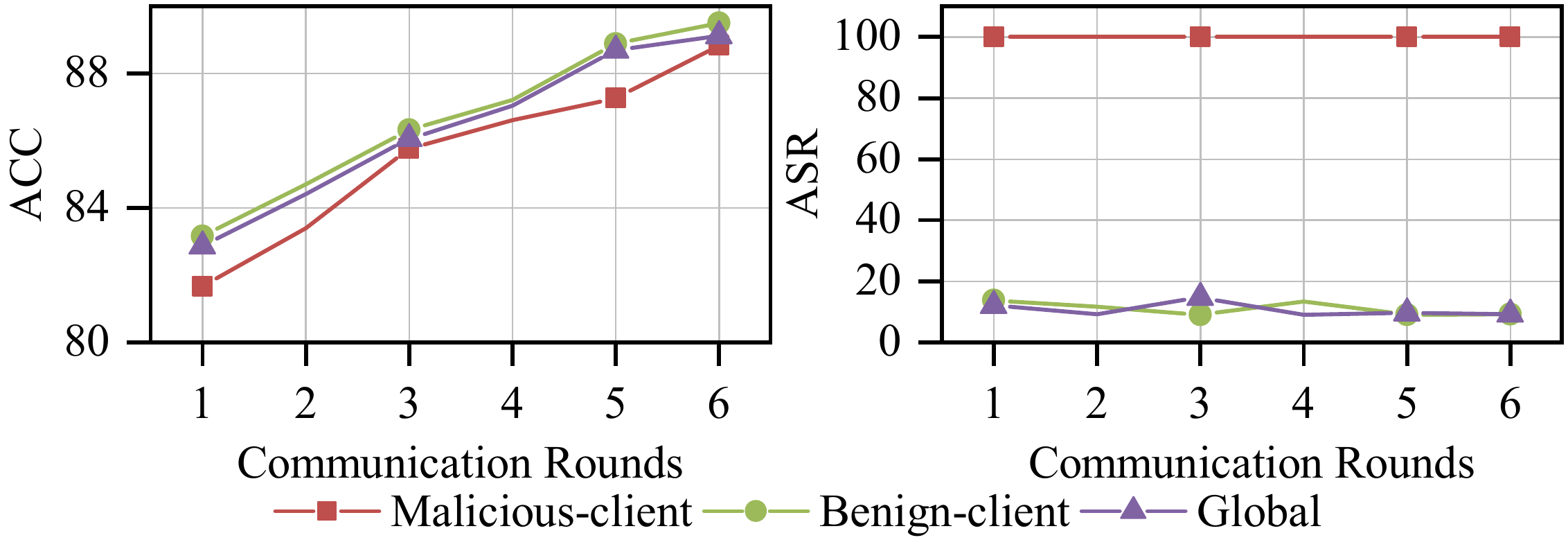} 
\caption{Local and global \emph{ACC (\%)} and \emph{ASR (\%)} with communication rounds on SST-2 task using BERT. The results are using FedPPT with IID setting and only one fixed client in ten clients is malicious. The benign client is selected randomly.}
\label{figasr}
\end{figure}

\begin{figure*}[t]
\centering
\includegraphics[width=2.0\columnwidth]{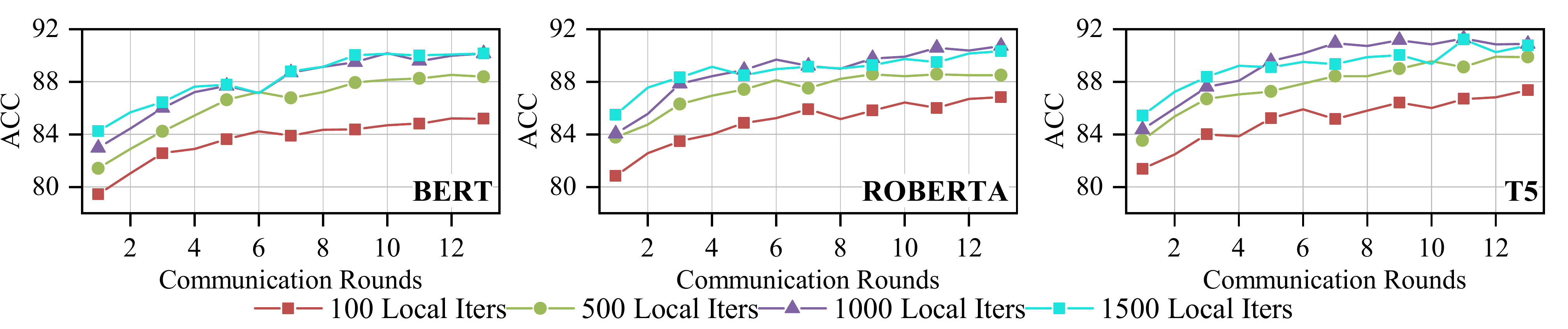} 
\caption{Global \emph{ACC (\%)} results with different local iterations on SST-2 task.}
\label{figit}
\end{figure*}

\subsection{Communication Rounds}
Fig. \ref{figacc} and Fig. \ref{figasr} show the local and global \emph{ACC} and \emph{ASR} in each round during training. As for \emph{ACC}, the training process of different settings is similar, and there is not a obvious decrease in Non-IID setting. The \emph{ACC} of local model in the first round have a rapid rise and pass it to the global model only in one single communication round. This proves that our proposed FedPrompt fits the poor data dependent prompt tuning well. As for \emph{ASR}, we can see that \emph{ASR} on the malicious client reaches 100\% only after one single local training round, but after aggregation, \emph{ASR} on benign client and global model remains a low level. We also find \emph{ASR} on benign client is close to global model in the previous round, consistent with the aggregation method. 

\subsection{Number of Local Iterations}
We study the effects of number of local iterations in each round. As we mentioned before, FedPrompt has relatively few trainable parameters that too many local iterations may lead to local over-fitting in FL while inadequate local iterations slow down the convergence. We conduct experiments on 100, 500, 1000 and 1500 local iterations, all of which are relatively small. As shown in Fig. \ref{figit}, 100 and 500 local iterations performs worse and 1500 iterations may lead to local over-fitting, which are harmful to obtain a excellent global model. Additional experiments expanded to 50 rounds suggest the performance of 100 and 500 local iterations still below others, while the other two could not get a further promotion.

\begin{table}[t]
\caption{Global \emph{ACC (\%)} results with different number of soft tokens on SST-2 task.}
\centering
\begin{tabular}{@{}ccccc@{}}
\toprule
Token Num & 1     & 5     & 10    & 20    \\ \midrule
\emph{ACC}       & 87.11 & 89.28 & 89.62 & 90.16 \\ \bottomrule
\end{tabular}

\label{tabtoken}
\end{table}

\begin{table}[!t]
\caption{Global \emph{ACC (\%)} results with and without LDP on SST-2 task.}
\centering
\begin{tabular}{@{}cccc@{}}
\toprule
Method                                                     & BERT  & ROBERTA & T5    \\ \midrule
\begin{tabular}[c]{@{}c@{}}FedPrompt w/o LDP\end{tabular} & 90.16 & 92.43   & 92.69 \\
\begin{tabular}[c]{@{}c@{}}FedPrompt w/ LDP\end{tabular}  & 85.73 & 86.88   & 86.14 \\ \bottomrule
\end{tabular}

\label{tabdp}
\end{table}

\begin{table}[!t]\small
\caption{Global \emph{ACC (\%)} and communication cost (M, million) with different prompt methods. Denote prompt tuning as method $\alpha$, P-tuning as $\beta$ and Prefix-Tuning as $\gamma$.}
\centering
\begin{tabular}{@{}c|cc|cc|cc@{}}
\toprule
\multirow{2}{*}{Method} & \multicolumn{2}{c|}{BERT} & \multicolumn{2}{c|}{ROBERTA} & \multicolumn{2}{c}{T5} \\
                        & \emph{ACC}         & Comm.       & \emph{ACC}          & Comm.         & \emph{ACC}       & Comm.      \\ \midrule
$\alpha$           & 90.16       & 0.016      & 92.43        & 0.016        & 92.69     & 0.015     \\
$\beta$                & 90.99           & 25.420           & 93.27            & 25.420             & 93.35         & 25.420          \\
$\gamma$           & -        & -           & -         & -             & 76.85         & 9.853          \\ \bottomrule
\end{tabular}%

\label{tabprompt}
\end{table}
\subsection{Number of Soft Tokens}
We tested the results under different numbers of soft tokens settings, and the results are consistent with those of prompt tuning under non-federal learning. As shown in Table \ref{tabtoken}, using more soft tokens will lead to better results. However, it also increases the communication cost under FL.

\subsection{FedPrompt with LDP}
As we mentioned before, there are hidden dangers to infer the origin private data by inverting gradients in FL, and LDP is an effective way to defense this attack. Also \cite{2021basu} has proved that the noise for larger model can damage the accuracy in differentially FL, so in Fedprompt the tiny prompt contributes to the use of LDP for privacy. We test on SST-2, clipping the gradients and then adding LaPlace noise on parameters. Table \ref{tabdp} shows that LDP protects the privacy with the cost of accuracy decreased by about 5\%.

\subsection{Prompt methods}
We also experiment on P-tuning and Prefix-Tuning (only supports T5 now\footnote{ https://github.com/thunlp/OpenPrompt}). Our experiments on SST-2 are shown in Table \ref{tabprompt}. It suggests that among the three prompt methods prompt tuning gets the best performance combining \emph{ACC} and communication cost. P-tuning has the best \emph{ACC} performance but quite a lot parameters, and Prefix-Tuning needs more modification to use.

\section{Further Improvement}
Though FedPrompt is robust to normal backdoor attack in our experiments, there are still special methods to backdoor FL. We plan to let the server check the mean and standard deviation of soft prompt parameters from each client, and find outliers to refuse before global aggregation. Adding noise after aggregation could also destroy the backdoor, with partial sacrifice on \emph{ACC}. We will carry on this research next.

\section{Conclusion}
In this work, we propose FedPrompt to use federated prompt tuning on decentralized data in a communication-efficient and privacy preserving way. We employ a split aggregation way that freezing extensive PLMs' parameters and only tuning and aggregating soft prompts. In this way we condense the communication cost to only 0.01\% compared to full-parameter fine-tuning, making many devices applicable for some scenarios with communication constraints. Experiments on both IID and Non-IID data distribution using three mainstream model demonstrate the accuracy of FedPrompt. We also use LDP to further protect the privacy, and it is necessary to further study the FL backdoor attack.




\bibliographystyle{IEEEtran}
\bibliography{IEEEabrv,refs}

\end{document}